\pdfoutput=1

\documentclass[11pt]{article}

\usepackage[]{naacl2021}

\usepackage{times}
\usepackage{latexsym}

\usepackage[T1]{fontenc}

\usepackage[utf8]{inputenc}

\usepackage{microtype}

\usepackage{xspace}

\usepackage{amsmath}
\usepackage{amsfonts}
\usepackage{amssymb}
\usepackage{algorithm,algpseudocode}
\usepackage{booktabs}
\usepackage{url}
\usepackage{graphicx}
\usepackage{comment}
\usepackage{makecell}
\usepackage{arydshln}
\usepackage{paralist}
\usepackage{soul}
\usepackage{algpseudocode}
\usepackage{varwidth}
\usepackage{subcaption}

\newcommand{\secref}[1]{\S\ref{#1}}

\newcommand{\ourmodel}{CLARE\xspace}
\newcommand{\mask}{\textsc{[Mask]}\xspace}

\newcommand{\interalia}[1]{\citep[\emph{inter alia}]{#1}}

\definecolor{mypurple}{RGB}{111,61,121}
\definecolor{myblue}{RGB}{46,88,180}
\definecolor{myred}{RGB}{181,68,106}
\definecolor{myyellow}{RGB}{204,143,55}

\newcommand{\replacecolor}[1]{{\color{myred}{\textit{#1}}}}
\newcommand{\insertcolor}[1]{{\color{myblue}{\textbf{#1}}}}
\newcommand{\mergecolor}[1]{{\color{myyellow}{\textsf{#1}}}}

\newcommand{\Replace}{\emph{Replace}\xspace}
\newcommand{\Insert}{\emph{Insert}\xspace}
\newcommand{\Merge}{\emph{Merge}\xspace}

\def\vx{{\mathbf{x}}}

\def\gA{{\mathcal{A}}}

\def\gV{{\mathcal{V}}}

\def\gZ{{\mathcal{Z}}}

\title{Contextualized Perturbation for Textual Adversarial Attack}

\author{Dianqi Li$^\spadesuit$ \quad 
Yizhe Zhang$^\diamondsuit$\quad 
Hao Peng$^\spadesuit$\quad
Liqun Chen$^\clubsuit$  \\
\textbf{Chris Brockett}$^\diamondsuit$\quad 
\textbf{Ming-Ting Sun}$^\spadesuit$ \quad 
\textbf{Bill Dolan}$^\diamondsuit$\\
$^\spadesuit$University of Washington \quad
$^\diamondsuit$Microsoft Research \quad 
$^\clubsuit$Duke University\\
{\tt \{dianqili, mts\}@uw.edu, hapeng@cs.uw.edu, liqun.chen@duke.edu} \\
{\tt \{Yizhe.Zhang, Chris.Brockett, billdol\}@microsoft.com}\\}

\begin{document}
\maketitle
\begin{abstract}
Adversarial examples expose the vulnerabilities of natural language processing (NLP) models, and can be used to evaluate and improve their robustness. Existing techniques of generating such examples are typically driven by local heuristic rules that are agnostic to the context, often resulting in unnatural and ungrammatical outputs. This paper presents \ourmodel, a \textbf{C}ontextua\textbf{L}ized \textbf{A}dversa\textbf{R}ial \textbf{E}xample generation model that produces fluent and grammatical outputs through a mask-then-infill procedure. \ourmodel builds on a pre-trained masked language model and modifies the inputs in a context-aware manner. We propose three contextualized perturbations, \Replace, \Insert and \Merge, that allow for generating outputs of varied lengths. \ourmodel can flexibly combine these perturbations and apply them at any position in the inputs, and is thus able to attack the victim model more effectively with fewer edits. Extensive experiments and human evaluation demonstrate that \ourmodel outperforms the baselines in terms of attack success rate, textual similarity, fluency and grammaticality. 
\end{abstract}

\section{Introduction}

Adversarial example generation for 
natural language processing (NLP) tasks aims to perturb input text to trigger errors in machine learning models,
while keeping the output close to the original. Besides exposing system
vulnerabilities and 
helping improve their robustness and security~\interalia{zhao2017generating,wallace2019universal,cheng2019robust,jia2019certified}, adversarial examples are also used to analyze and interpret the models' decisions~\citep{jia2017adversarial,ribeiro2018semantically}.

Generating adversarial examples for NLP tasks can be challenging, in part due to the discrete nature of natural language text. 
Most recent efforts have explored heuristic rules, such as replacing tokens with their synonyms~\interalia{samanta2017towards,liang2017deep,alzantot2018generating,ren2019generating,jin2019bert}. 
Despite some empirical success, rule-based methods 
are agnostic to context, limiting their
ability to produce natural, fluent, and grammatical outputs~\interalia{wang2019natural,kurita2020weight}.


\begin{figure}[t]
\centering
\includegraphics[width=1.0\linewidth]{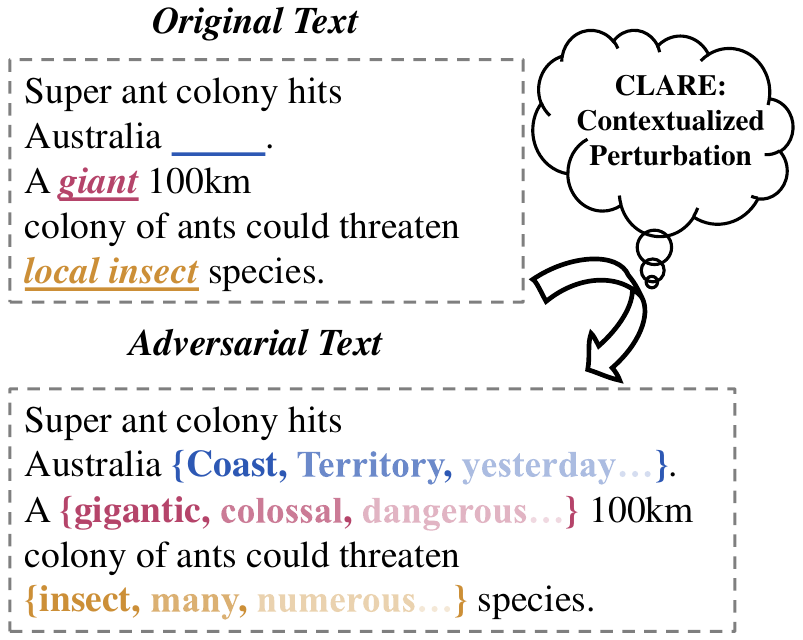}
\caption{Illustration of \ourmodel. Through a mask-then-infill procedure, the model generates the adversarial text with three contextualized perturbations: {\color{myred}{\textbf{\Replace}}}, {\color{myblue}{\textbf{\Insert}}} and {\color{myyellow}{\textbf{\Merge}}}. A mask is indicated by ``\underline{\hspace{5mm}}''. The degree of fade corresponds to the (decreasing) priority of the infill tokens.}

\label{fig:clare}
\end{figure}

This work presents \ourmodel, a \textbf{C}ontextua\textbf{L}ized \textbf{A}dversa\textbf{R}ial \textbf{E}xample generation model for text. 
\ourmodel perturbs the input with a mask-then-infill procedure:
it first detects the vulnerabilities of a model and deploys masks to the inputs to indicate missing text, 
then plugs in an alternative using a pretrained masked language model (e.g., RoBERTa;~\citealp{liu2019roberta}).
\ourmodel features three contextualized perturbations: \Replace, \Insert and \Merge, 
which respectively replace a token, insert a new one, and merge a bigram (Figure~\ref{fig:clare}). 
As a result, it can generate outputs of varied lengths, 
in contrast to token replacement based methods that are limited 
to outputs of the same lengths as the inputs \citep{alzantot2018generating,ren2019generating,jin2019bert}. 
Further, \ourmodel searches over a wider range of attack strategies,
and is thus able to attack the victim model more effectively with fewer edits. 
Building on a masked language model,
\ourmodel maximally preserves textual similarity, fluency, and grammaticality of the outputs. 



We evaluate \ourmodel on text classification, natural language inference, and sentence paraphrase tasks,
by attacking finetuned BERT models~\citep{devlin2019bert}. 
Extensive experiments and human evaluation results show that 
\ourmodel outperforms baselines in terms of attack success rate, textual similarity, fluency, and grammaticality,
and strikes a better balance between attack success rate and preserving input-output similarity. 
Our analysis further suggests that the \ourmodel can be used to improve the robustness of the downstream models,
and improve their accuracy when the available training data is limited. We release our code and models at \url{https://github.com/cookielee77/CLARE}.

\section{\ourmodel}
At a high level, \ourmodel 
applies a sequence of contextualized perturbation actions to the input. 
Each can be seen as a \emph{local} mask-then-infill procedure:
it first applies a mask to the input around a given position,
and then fills it in using a pretrained masked language model~(\S\ref{sec:operations}).
To produce the output, \ourmodel scores and descendingly ranks the actions, which are then iteratively 
applied to the input~(\S\ref{sec:ranking}). 
We begin with a brief background review and laying out of necessary notation.
\paragraph{Background.}
Adversarial example generation centers around a \textbf{victim} model $f$, which we assume is a text classifier.
We focus on the black-box setting,
allowing access to $f$'s outputs but \emph{not} its configurations such as parameters.
Given an input sequence $\vx=x_1x_2\dots x_n$ and its label $y$ (assume $f(\vx)=y$),
an \textbf{adversarial example} $\vx^\prime$ is supposed to modify $\vx$ to trigger an error in the victim model: 
$f(\vx^\prime) \neq f(\vx)$.
At the same time, textual modifications should be minimal,
such that $\vx^\prime$ is close to $\vx$ 
and the human predictions on $\vx^\prime$ stay the same.\footnotemark

\footnotetext{
    In computer vision applications, 
    minor perturbations to continuous pixels can be barely perceptible to humans, 
    thus it can be hard for one to distinguish $\vx$ and $\vx^\prime$~\citep{goodfellow2014explaining}.
    It is not the case for text, however, since
    changes to the discrete tokens are more likely to be noticed by humans.
}
This is achieved by requiring the similarity between
$\mathbf{x}^\prime$ and $\vx$ to be larger than a 
threshold: $\operatorname{sim}(\mathbf{x}^\prime,\mathbf{x}) > \ell$.
A common choice of $\operatorname{sim}(\boldsymbol{\cdot}, \boldsymbol{\cdot})$ is to encode sentences using neural networks,
and calculate their cosine similarity in the embedding space~\citep{jin2019bert}.

\subsection{Masking and Contextualized Infilling}\label{sec:operations}
At a given position of the input sequence,
\ourmodel can execute three perturbation actions: 
\Replace, \Insert, and \Merge,
which we introduce in this section.
These apply masks at the given position with different strategies,
and then fill in the missing text based on the unmasked context.

\paragraph{\Replace: } 
A \Replace action substitutes the token at a given position $i$ with an alternative (e.g., changing ``\emph{fantastic}'' to ``\emph{amazing}'' in ``The movie is \emph{fantastic}.''). 
It first replaces $x_i$
with a mask,
and then selects a token $z$ from a candidate set $\gZ$ to fill in:
\begin{align*}
    \widetilde{\vx} &= x_1\dots x_{i-1}\ \mask \ x_{i+1}\dots x_n,\\
    \operatorname{replace}\left(\mathbf{x}, i\right) &= x_1\dots x_{i-1}\ z \ x_{i+1}\dots x_n.
\end{align*}
For clarity, we denote $\operatorname{replace}\left(\mathbf{x}, i\right)$ by $\widetilde{\vx}_z$.
To produce an adversarial example, 
\begin{compactitem}
\item $z$ should fit into the unmasked context;
\item $\widetilde{\vx}_z$ should be similar to $\vx$;
\item $\widetilde{\vx}_z$ should trigger an error in $f$.
\end{compactitem}
These can be achieved by selecting a $z$ such that
\begin{compactitem}
\item $z$ receives a high probability from a masked language model: $p_{\text{MLM}}(z\mid \widetilde{\vx}) > k$;
\item $\widetilde{\vx}_z$ is similar to $\vx$: $\operatorname{sim}(\vx, \widetilde{\vx}_z) > \ell$;
\item $f$ predicts low probability for the gold label given $\widetilde{\vx}_z$, i.e., $p_f(y\mid \widetilde{\vx}_z)$ is small.
\end{compactitem}
$p_{\text{MLM}}$ denotes a pretrained masked language model (e.g., RoBERTa;~\citealp{liu2019roberta}). 
Using higher $k$, $\ell$ thresholds produces outputs that are more fluent and closer to the original. However, this can undermine the success rate of the attack. We choose $k$, $\ell$ to trade-off between these two aspects. \footnote{
    $k$ and $\ell$ are empirically set as $5\times 10^{-3}$ and $0.7$, respectively.
    This also reduces the computation overhead:
    in our experiments $|\gZ|$ is $42$ on average, much smaller than the vocabulary size ($|\gV|=50,265$).
} 

The first two requirements can be met by
the construction of the candidate set: $\gZ=$
\begin{align*}
    \left\{z^\prime\in\gV \mid p_{\text{MLM}}(z^\prime\mid \widetilde{\vx}) > k, \operatorname{sim}(\vx, \widetilde{\vx}_{z^\prime}) > \ell\right\}.
\end{align*}
$\gV$ is the vocabulary of the masked language model.
To meet the third, we select from $\gZ$ the token that, if filled in,
will cause most ``confusion'' to $f$:
\begin{align}
    z = \operatorname*{arg\,min}_{z^\prime \in \gZ} p_f(y\mid \widetilde{\vx}_{z^\prime}).
    \label{eq:eq1}
\end{align}

The \Insert and \Merge actions differ from \Replace
in terms of masking strategies. 
The alternative token $z$ is selected analogously to that in a \Replace action.

\paragraph{\Insert: } 
This aims to add extra information to the input (e.g., changing ``I recommend ...'' to ``I \emph{highly} recommend ...'').
It inserts a mask after $x_i$ and then fills it.
Slightly overloading the notations, 
\begin{align*}
    \widetilde{\vx} &= x_1\dots x_i\ \mask \ x_{i+1}\dots x_n,\\
    \operatorname{insert}\left(\mathbf{x}, i\right) &= x_1\dots x_{i}\ z \ x_{i+1}\dots x_n.
\end{align*}
This increases the sequence length by 1.

\paragraph{\Merge: } 
This masks out a bigram $x_i x_{i+1}$
with \emph{a single} mask and then fills it, reducing the sequence length by 1:
\begin{align*}
    \widetilde{\vx} &= x_1\dots x_{i-1}\ \mask \ x_{i+2}\dots x_n,\\
    \operatorname{merge}\left(\mathbf{x}, i\right) &= x_1\dots x_{i-1}\ z\ x_{i+2}\dots x_n.
\end{align*}
$z$ can be the same as one of the masked tokens (e.g., masking out ``New York'' and then filling in``York''). This can be seen as deleting a token from the input.

For \Insert and \Merge, $z$ is chosen in the same manner as replace action.
\footnote{
    A perturbation will not be considered if its candidate token set is empty.
}

In sum, at each position $i$ of an input sequence,
\ourmodel first:
$(i)$ replaces $x_i$ with a mask;
$(ii)$ or inserts a mask after $x_i$;
$(iii)$ or merges $x_i x_{i+1}$ into a mask.
Then a set of candidate tokens is constructed
with a masked language model and a textual similarity function; the token minimizing the gold label's probability is chosen as the alternative token. The combination of these three operations enables conversion between any two sequences. 

\ourmodel first constructs the local actions for all positions in parallel, 
i.e., the actions at position $i$ do not affect those at other positions.
Then, to produce the adversarial example,
\ourmodel gathers the local actions and selects an order to execute them.

\subsection{Sequentially Applying the Perturbations}\label{sec:ranking}
Given an input pair $(\vx, y)$,
let $n$ denote the length of $\vx$.
\ourmodel chooses from $3n$ actions to produce the output:
3 actions for each position, assuming the candidate token sets are not empty.
We aim to generate an adversarial example with minimum modifications to the input.
To achieve this, we iteratively apply the actions,
and first select those minimizing the probability of outputting the gold label $y$ from $f$.

Each action is associated with a score, measuring how likely it can ``confuse'' $f$:
denote by $a(\vx)$ the output of applying action $a$ to $\vx$.
The score is then the negative probability of predicting the gold label from $f$, using $a(\vx)$ as the input:
\begin{align*}\label{eq:action_score}
    s_{(\vx, y)}(a) = -p_f\bigl(y\mid a(\vx)\bigr).
\end{align*}
\emph{Only one} of the three actions can be applied at each position, and we select the one with the highest score.
This constraint aims to avoid multiple modifications around the same position, e.g.,
merging ``New York'' into ``Seattle'' and then replacing it with ``Boston''.

Actions are iteratively applied to the input, until an adversarial example is found 
or a limit of actions $T$ is reached.
Each step selects the highest-scoring action from the remaining ones.
Algorithm~\ref{alg:oscar} summarizes the above procedure.\footnote{
    \Insert and \Merge actions change the text length.
    When any of them is applied, we accordingly change the text indices of affected actions remaining in $\gA$.
}

\begin{algorithm}[t]
\centering
\caption{Adversarial Attack by \ourmodel}
\label{alg:oscar}
\begin{algorithmic}[1]
\State{\bfseries Input:} Text-label pair $(\vx, y)$;  Victim model $f$
\State{\bfseries Output:} An adversarial example
\State{\bfseries Initialization:} $\vx^{(0)} = \vx$
\State{$\gA\leftarrow \varnothing$}
\For {$1 \leq i \leq \lvert\vx\rvert$}
    \State \begin{varwidth}[t]{\linewidth}
     $a \leftarrow$ highest-scoring action from $\{$\par
    \hskip\algorithmicindent$\operatorname{replace}(\vx, i)$, $\operatorname{insert}(\vx, i), \operatorname{merge}(\vx, i)\}$
    \end{varwidth}
    \State $\gA\leftarrow \gA \bigcup \{a\}$
\EndFor

\For{$1 \leq t \leq T$}
    \State $a\leftarrow$ highest-scoring action from $\gA$
    \State $\gA\leftarrow \gA \setminus \{a\}$
    \State{$\textbf{x}^{(t)} \leftarrow$ Apply $a$ on $\textbf{x}^{(t-1)}$}
    \If{$f(\vx^{(t)}){\neq}y$} {\Return $\vx^{(t)}$}
    \EndIf
\EndFor
\State \Return \textsc{None}

\end{algorithmic}
\end{algorithm}

\paragraph{Discussion.}
A key technique of \ourmodel is the local mask-then-infill perturbation. Compared with existing context-agnostic replacement approaches~\interalia{alzantot2018generating,jin2019bert,ren2019generating}, contextualized infilling produces more fluent and grammatical outputs. Generating adversarial examples with masked language models is also explored by concurrent work BERTAttack~\citep{li2020bert} and BAE~\citep{garg2020bae}.\footnote{
Both \citet{li2020bert} and \citet{garg2020bae} are published concurrently to an initial report of this work.
}
\begin{compactitem}
\item
BERTAttack only replaces tokens and thus 
can only produce outputs of the same lengths as the inputs.
This is analogous with a \ourmodel model
with the \Replace action only.
BAE entangles replacing and inserting tokens:
it inserts \emph{only} at positions neighboring a replaced token,
limiting its attacking capability.
Departing from both, \ourmodel uses three different perturbations (\Replace, \Insert and \Merge),
each allowing efficient attacking against \emph{any} position of the input,
and can produce outputs of varied lengths. 
As we will show in the experiments (\secref{sec:exp_res}), \ourmodel outperforms both these methods.

\item
When selecting the attack positions,
neither BERTAttack or BAE
takes into account the tokens to be infilled,
whereas \ourmodel does.
This results in better adversarial attack performance according to our ablation study~(\S\ref{ana:abl}).

\item \ourmodel demonstrates the advantage of using RoBERTa over BERT, which was used in the concurent works~(\S\ref{ana:abl}).
\end{compactitem}

\section{Experiments}\label{sec:exp}
We evaluate \ourmodel on text classification, natural language inference, and sentence paraphrase tasks. We begin by describing the implementation details of \ourmodel and the baselines~(\S\ref{sec:setup}).
\S\ref{sec:data} introduces the experimental datasets and the evaluation metrics;
the results are summarized in \S\ref{sec:exp_res}.

\subsection{Setup}\label{sec:setup}
\begin{compactitem}
\item We experiment with a distilled version of RoBERTa (RoBERTa$_{\text{distill}}$; \citealp{sanh2019distilbert}) as the masked language model for contextualized infilling.
We also compare to base sized RoBERTa (RoBERTa$_{\text{base}}$; \citealp{liu2019roberta})
and base sized BERT (BERT$_{\text{base}}$; \citealp{devlin2019bert}) in the ablation study (\secref{ana:abl}).
\item The similarity function builds on the universal sentence encoder (USE;~\citealp{cer2018universal}).
\item The victim model is an MLP classifier on top of BERT$_{\text{base}}$. 
It takes as input the first token's contextualized representation. We finetune BERT when training the victim model.
\end{compactitem}

\paragraph{Baselines.}
We compare \ourmodel with recent state-of-the-art word-level black-box adversarial attack models, including:
\begin{compactitem}
\item{\bf TextFooler}: a state-of-the-art model 
by \citet{jin2019bert}.
This replaces tokens with their synonyms derived from counter-fitting word embeddings~\citep{mrkvsic2016counter},
and uses the same text similarity function as our work.

\item{\bf TextFooler+LM}:
an improved variant of TextFooler we implemented based on \citet{alzantot2018generating} and
\citet{cheng2019robust}.
This inherits token replacement from TextFooler, but
uses an additional small sized GPT-2 language model~\citep{radford2019language} to filter out those candidate tokens that do not fit in the context with calculated perplexity.

\item{\bf BERTAttack}:
a mask-then-infill approach by~\citet{li2020bert}. It greedily replaces tokens with the predictions from BERT. BAE is not listed as it has a similar performance as BERTAttack \citep{garg2020bae}.
\end{compactitem}

We use the open source implementation of the above baselines provided by the authors. More details are included in Appendix~\secref{sec:app_details}.



\begin{table}[t]
\centering
\setlength{\tabcolsep}{4.0pt}
\small{
\begin{tabular}{@{} lccrrr @{}}
\toprule

\textbf{Dataset}  & \textbf{Avg. Length} & \textbf{\# Classes} &
\textbf{Train} & \textbf{Test} & \textbf{Acc}
\\\midrule

Yelp & 130 & 2 & 560K & 38K & 95.9\% \\
AG News & 46 & 4 & 120K & 7.6K & 95.0\%  \\
\midrule

MNLI\footnotemark & 23/11 & 3 & 392K & 9.8K & 84.3\%\\
QNLI & 11/31 & 2 & 105K & 5.4K & 91.4\%
\\\bottomrule
\end{tabular}
}
\caption{Some statistics of datasets. 
The last column indicates the victim model's accuracy on the original test set \emph{without} adversarial attack.}
\label{tab:dataset}
\end{table}
\footnotetext{We only examine the performance on the matched set, since the mismatched set is easier to attack.}

\begin{table*}[t]
\centering
\setlength{\tabcolsep}{6pt}
\small{
\begin{tabular}{@{} l rrrrr m{3pt} rrrrr @{}}
\toprule
{} & \multicolumn{5}{c}{\textbf{Yelp} (PPL = 51.5)} && \multicolumn{5}{c}{\textbf{AG News} (PPL = 62.8)}
\\\midrule
\textbf{Model}  &
\textbf{A-rate}$\uparrow$ & \textbf{Mod}$\downarrow$ & \textbf{PPL}$\downarrow$ & \textbf{GErr}$\downarrow$ & \textbf{Sim}$\uparrow$ & &
\textbf{A-rate}$\uparrow$ & \textbf{Mod}$\downarrow$ & \textbf{PPL}$\downarrow$ & \textbf{GErr}$\downarrow$ & \textbf{Sim}$\uparrow$
\\\midrule
 TextFooler  &
 77.0 &  16.6 &  163.3 &  1.23 &  0.70 &&
 56.1 &  23.3 &  331.3 &  1.43 &  0.69
 \\
 \quad + LM  &
 34.0 &  17.4 &  90.0 &  1.21 &  0.73 &&
 23.1 &  21.9 &  144.6 &  1.07 &  0.74
 \\
  BERTAttack  &
 71.8 &  10.7 &  90.8 &  0.27 &  0.72 &&
 63.4 &  7.9 &  90.6 &  0.25 &  0.71
 \\
 \midrule
 \textsc{Clare}  &
\textbf{79.7} &  \textbf{10.3} &  \textbf{83.5} &  \textbf{0.25} &  \textbf{0.78} &&
\textbf{79.1} &  \textbf{6.1} &  \textbf{86.0} &  \textbf{0.17} &  \textbf{0.76}
\\\midrule\midrule

{} & \multicolumn{5}{c}{\textbf{MNLI} (PPL = 60.9)} && \multicolumn{5}{c}{\textbf{QNLI} (PPL = 46.0)}
\\\midrule
\textbf{Model}  &
\textbf{A-rate}$\uparrow$ & \textbf{Mod}$\downarrow$ & \textbf{PPL}$\downarrow$ & \textbf{GErr}$\downarrow$ & \textbf{Sim}$\uparrow$ & &
\textbf{A-rate}$\uparrow$ & \textbf{Mod}$\downarrow$ & \textbf{PPL}$\downarrow$ & \textbf{GErr}$\downarrow$ & \textbf{Sim}$\uparrow$
\\\midrule
 TextFooler  &
 59.8 &  13.8 &  161.5 &  0.63 &  0.73 &&
 57.8 &  16.9 &  164.4 &  0.62 &  0.72
 \\
 \quad + LM  &
 32.3 &  12.4 &  91.9 &  0.50 &  0.77 &&
29.2 &  17.3 &  85.0 &  0.42 &  0.75
 \\
  BERTAttack  &
 82.7 &  8.4 &  86.7 &  0.04 &  0.77 &&
 76.7 &  13.3 &  86.5 &  0.03 &  0.73
 \\
 \midrule
 \textsc{Clare}  &
\textbf{88.1} &  \textbf{7.5} &  \textbf{82.7} &  \textbf{0.02} &  \textbf{0.82} &&
\textbf{83.8} &  \textbf{11.8} &  \textbf{76.7} &  \textbf{0.01} &  \textbf{0.78}
\\\bottomrule
\end{tabular}
}
\caption{Adversarial example generation performance in
attack success rate (A-rate),
modification rate (Mod),
perplexity (PPL),
number of increased grammar errors (GErr),
and textual similarity (Sim).
The perplexity of the original inputs is indicated in parentheses for each dataset.
Bold font indicates the best performance for each metric. All numbers are reported on 1000 test instances. $\uparrow$ ($\downarrow$) represents that the higher (lower) the better.
}
\label{tab:res}
\end{table*}

\subsection{Datasets and Evaluation}\label{sec:data}
\paragraph{Datasets.}
We evaluate \ourmodel 
with the following datasets: 

\begin{compactitem}
\item \textbf{Yelp Reviews}~\citep{zhang2015character}: a binary sentiment classification dataset based on restaurant reviews. 

\item \textbf{AG News}~\citep{zhang2015character}: a collection of news articles with four categories: \emph{World}, \emph{Sports}, \emph{Business} and \emph{Science \& Technology}.

\item \textbf{MNLI}~\citep{williams2018broad}: a natural language inference
dataset. Each instance consists of a premise-hypothesis pair, and the model is supposed to determine the relation
between them from a label set of \emph{entailment}, \emph{neutral}, and \emph{contradiction}.
It covers text from a variety of domains.

\item \textbf{QNLI}~\citep{wang2019glue}: a binary classification dataset converted from the Stanford question answering dataset~\citep{rajpurkar2016squad}. 
The task is to determine whether the context contains
the answer to a question. It is mainly based on English Wikipedia articles.
\end{compactitem}

Table~\ref{tab:dataset} summarizes some statistics of the datasets. 
In addition to the above four datasets, we experiment with 
DBpedia ontology dataset~\citep{zhang2015character}, 
Stanford sentiment treebank \citep{socher2013recursive}, 
Microsoft Research Paraphrase Corpus \citep{dolan2005automatically},
and Quora Question Pairs from the GLUE benchmark. 
The results on these datasets are summarized in Appendix~\ref{sec:app_res}.

Following previous practice~\citep{alzantot2018generating}, 
we fine-tune \ourmodel on training data,
and evaluate with 1,000 randomly sampled test instances of lengths $\leq100$.
In the sentence-pair tasks (e.g., MNLI, QNLI), we attack the longer sentence excluding the tokens that appear in both.

\paragraph{Evaluation metrics.}
We follow previous works \citep{jin2019bert,morris2020textattack},
and evaluate the models with the following automatic metrics:
\begin{compactitem}
\item {\bf Attack success rate (A-rate)}: the percentage of adversarial examples that can successfully attack the victim model. 

\item {\bf Modification rate (Mod)}: the percentage of modified tokens.
Each \Replace or \Insert action accounts for one token modified;
a \Merge action is considered modifying one token
if one of the two merged tokens is kept (e.g., merging bigram $a b$ into $a$),
and two otherwise (e.g., merging bigram $a b$ into $c$).

\item{\bf Perplexity (PPL)}: a metric used to evaluate the \textit{fluency} of adversaries~\citep{kann2018sentence,zang2020word}. The perplexity is calculated using small sized GPT-2 with a 50K-sized vocabulary~\citep{radford2019language}.

\item{\bf Grammar error (GErr)}: the absolute number of increased grammatical errors 
in the successful adversarial example, compared to the original text. Following~\citep{zang2020word,morris2020reevaluating}, we
calculate this by the LanguageTool~\citep{naber2003rule}.\footnote{\url{https://www.languagetool.org/}}

\item{\bf Textual similarity (Sim)}: the cosine similarity between the input and its adversary. Following \citep{jin2019bert,morris2020reevaluating}, we calculate this using the universal sentence encoder (USE; \citealp{cer2018universal}). 
\end{compactitem}

The last four metrics are averaged across those adversarial examples that successfully attack the victim model.

\subsection{Results} \label{sec:exp_res}

Table~\ref{tab:res} summarizes the results.
Overall \ourmodel achieves the best performance on all metrics consistently across different datasets.
Notably, \ourmodel outperforms BERTAttack, the strongest baseline,
by a more than 5.4\% attack success rate with \emph{fewer} average modifications to the text.
We attribute this to \ourmodel's flexible attack strategies obtained by combining three different perturbations at any position.
Interestingly, using contextualized embeddings does \emph{not}
appear to guarantee better fluency:
despite fewer modifications to the text,
BERTAttack achieves similar perplexity to language-model-augmented TextFooler on three out of the four datasets,
while \ourmodel consistently outperforms both.
In terms of grammatical errors,
contextualized models (\ourmodel and BERTAttack) 
are substantially better than the others, with \ourmodel performing the best. In terms of similarity, \ourmodel outperforms all baselines by more than 0.02,
a larger gap than BERTAttack's improvements over TextFooler variants.
We observe similar trends on other datasets in Appendix~\ref{sec:app_res}.



Figure~\ref{fig:similarity-trade-off} compares trade-off curves 
between attack success rate and textual similarity.
We tune the thresholds for constructing the candidate token sets,
and plot textual similarity against the attack success rate.
\ourmodel strikes the best balance, showing a clear advantage in success rate with least similarity drop.
We observe similar trends for attack success rate and perplexity trade off. 

\begin{figure}[t]
\centering
\begin{minipage}{0.494\linewidth}
\includegraphics[width=1.0\linewidth]{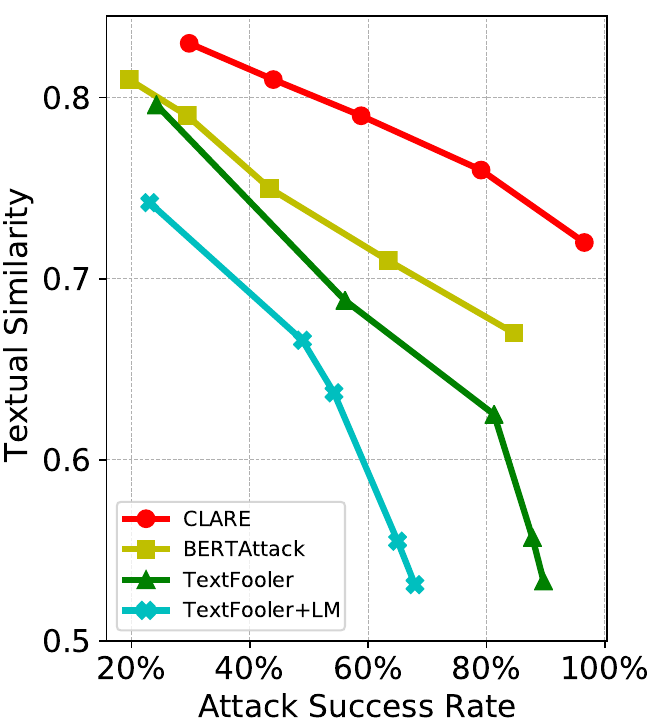}
\end{minipage}
\begin{minipage}{0.494\linewidth}
\includegraphics[width=1.0\linewidth]{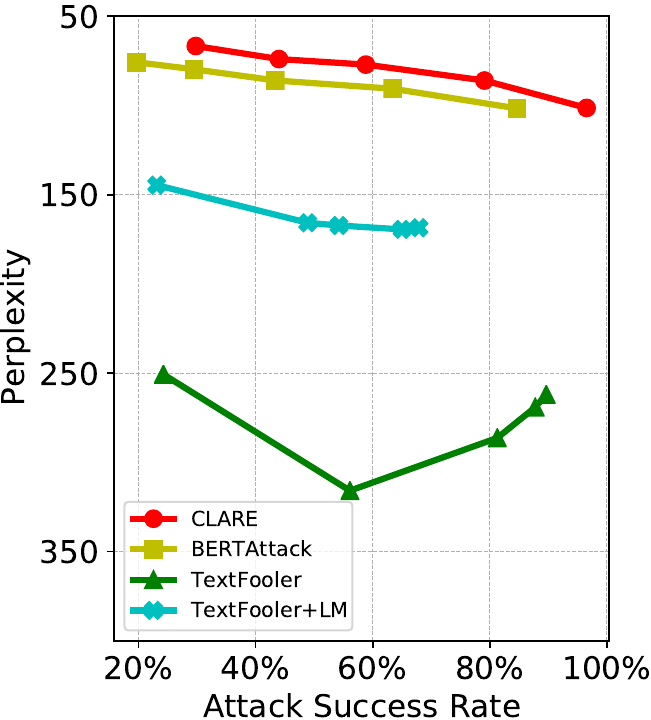}
\end{minipage}

\caption{\textbf{Left}: Attack success rate and textual similarity trade-off curves (\textit{both higher the better}). \textbf{Right}: Attack success rate (\textit{higher the better}) and perplexity (\textit{lower the better}) trade-off curve.
The larger area under the two curves indicates the better trade-off between two metrics.
}
\label{fig:similarity-trade-off}
\end{figure}

\paragraph{Human evaluation.} 
We further conduct human evaluation on the AG News dataset.
We randomly sample 300 instances which both \ourmodel and TextFooler successfully attack.
For each input, we pair the adversarial examples from the two models,
and present them to crowd-sourced judges along with the original input and the gold label.
We ask them which they prefer with a neutral option in terms of
(1) having a meaning that is closer to the original input (similarity), 
and (2) being more fluent and grammatical (fluency and grammaticality).
Additionally, we ask the judges to annotate adversarial examples,
and compare their annotations against the gold labels (label consistency). We collect 5 responses for each pair on every evaluated aspect. 
Further details are in Appendix~\ref{sec:app_human_eval}. 

As shown in Table~\ref{tab:human},
\ourmodel has a significant advantage over TextFooler:
in terms of similarity 56\% responses prefer \ourmodel,
while 16\% prefer TextFooler. 
The trend is similar for fluency \& grammaticality (42\% vs. 9\%). 
This observation is consistent with results from automatic metrics. On label consistency,
\ourmodel slightly underperforms TextFooler at 68\% with a 95\% condidence interval (CI) $(66\%, 70\%)$,
versus 70\% with a 95\% CI $(68\%, 73\%)$. We attribute this to an inherent overlap of some categories in the AG News dataset, e.g., \emph{Science \& Technology} and \emph{Business}, as evidenced by a 71\% label consistency for original inputs.

Closing this section,
Table~\ref{tab:samples} compares the adversarial examples generated by
TextFooler and \ourmodel. 
More samples are listed in Appendix~\ref{sec:app_samples}.


\begin{table}[t]
\centering
\setlength{\tabcolsep}{4pt}
\small{
\begin{tabular}{@{} l ccc @{}}
\toprule


Metric & \ourmodel & Neutral & TextFooler
\\\midrule
Similarity & 56.1$_{\pm 2.5}$ & 28.1 & 15.8$_{\pm 2.1}$ \\
Fluency\&Grammaticality & 42.5$_{\pm 2.5}$ & 48.6 & \phantom{1}8.9$_{\pm 1.5}$ \\
Label Consistency & 68.0$_{\pm 2.4}$ & - & 70.1$_{\pm 2.5}$
\\\bottomrule

\end{tabular}
}
\caption{Human evaluation performance in percentage on the AG News dataset. 
    $\pm$ indicates confidence intervals with a 95\% confidence level.
}
\label{tab:human}
\end{table}

\begin{table}[t]
\small
\begin{tabular}{@{} ll @{}}
 \noalign{\smallskip}\Xhline{3\arrayrulewidth}\noalign{\smallskip}
 \begin{minipage}{0.5in}
  \textbf{AG} \\
  (Sci\&Tech)
\end{minipage}
&  
\begin{minipage}{2.35in}
  Sprint Corp. is in talks with Qualcomm Inc. about using a network the chipmaker is building to deliver live television to Sprint mobile phone customers.
\end{minipage}
 \\\noalign{\smallskip}\hdashline\noalign{\smallskip}

\begin{minipage}{0.5in}
  TextFooler\\
  (Business)
\end{minipage}
&  
\begin{minipage}{2.35in}
  Sprint \replacecolor{Corps}. is in talks with Qualcomm Inc. about \replacecolor{operated} a network the chipmaker is \replacecolor{consolidation} to \replacecolor{doing} \replacecolor{viva} television to Sprint mobile phone customers.
\end{minipage}
 \\\noalign{\smallskip}\hdashline\noalign{\smallskip}
 
 \begin{minipage}{0.5in}
  \textsc{Clare}\\
  (Business)
\end{minipage}
&  
\begin{minipage}{2.35in}
  Sprint Corp. is in talks with Qualcomm Inc. about using a network \mergecolor{Qualcomm} is building to deliver \replacecolor{cable} television to Sprint mobile phone customers.
\end{minipage}
 \\\noalign{\smallskip}\Xhline{3\arrayrulewidth}\noalign{\smallskip}
 
 
 \begin{minipage}{0.5in}
  \textbf{MNLI} \\
  (Neutral)
\end{minipage}
&  
\begin{minipage}{2.35in}
  \emph{Premise}: Let me try it. She began snapping her fingers   and saying the word eagerly, but nothing happened.\\
  \emph{Hypothesis}:  She became frustrated when the spell didn't work.
\end{minipage}
 \\\noalign{\smallskip}\hdashline\noalign{\smallskip}

\begin{minipage}{0.5in}
  TextFooler\\
  (Contra-\\\phantom{$^\dagger$}diction)
\end{minipage}
&  
\begin{minipage}{2.35in}
  \emph{Premise}: \replacecolor{Authorisation} me \replacecolor{attempting} it. She \replacecolor{triggered} \replacecolor{flapping} her \replacecolor{pinkies} and \replacecolor{said} the word eagerly, but nothing \replacecolor{arisen}.\\
  \emph{Hypothesis}:  She became frustrated when the spell didn't work.
\end{minipage}
 \\\noalign{\smallskip}\hdashline\noalign{\smallskip}
 
 \begin{minipage}{0.5in}
  \textsc{Clare}\\
  (Contra-\\\phantom{$^\dagger$}diction)
\end{minipage}
&  
\begin{minipage}{2.35in}
  \emph{Premise}: Let me try it. She began snapping her fingers   and saying the word eagerly, but nothing \insertcolor{unexpected} happened.\\
  \emph{Hypothesis}:  She became frustrated when the spell didn't work.
\end{minipage}
 \\\noalign{\smallskip}\Xhline{3\arrayrulewidth}\noalign{\smallskip}
 
\end{tabular}
\caption{Adversarial examples produced by different models. The gold label of the original is shown below the (bolded) dataset name. \replacecolor{\textbf{Replace}}, \insertcolor{\textbf{Insert}} and \mergecolor{\textbf{Merge}} are highlighted in \replacecolor{italic red}, \insertcolor{bold blue} and \mergecolor{sans serif yellow}, respectively. (Best viewed in color).}
\label{tab:samples}
\end{table}

\section{Analysis}
This section first conducts an ablation study~(\S\ref{ana:abl}).
We then explore \ourmodel's potential to be used to improve downstream models' robustness and accuracy in
\S\ref{ana:adv_training}.
In \S\ref{ana:pos}, we empirically observe that \ourmodel tends to attack noun and noun phrases.

\subsection{Ablation Study} \label{ana:abl}

\begin{table}[t]
\centering
\setlength{\tabcolsep}{4.5pt}
\small{
\begin{tabular}{@{} l@{\hskip 0.1mm}rrrrr @{}}
\toprule

\textbf{Module}  &
\textbf{A-rate}$\uparrow$ & \textbf{Mod}$\downarrow$ & \textbf{PPL}$\downarrow$ & \textbf{GErr}$\downarrow$ & \textbf{Sim}$\uparrow$ \\
\midrule

\textsc{Clare} & 79.1 & \textbf{6.1} & \textbf{86.0} & 0.17 & 0.76 \\
\midrule
\textsc{MergeOnly}\footnotemark & 47.2 & 6.2 & 95.3 & \textbf{0.08} & \textbf{0.79} \\
\textsc{InsertOnly} & 68.1 & 7.2 & 93.1 & 0.23 & 0.74 \\
\textsc{ReplaceOnly} & 66.7 & 7.7 & 85.6 & 0.10 & 0.72 \\
\quad BERTAttack & 63.4 &  7.9 &  90.6 &  0.25 &  0.71 \\
\midrule
\emph{w/o} $\operatorname{sim} > \ell$ & 82.4 & 6.9 & 86.8 & 0.13 & 0.70 \\
\emph{w/o} $p_{\text{MLM}} > k$ & \textbf{95.7} & 6.8 & 162.8 & 0.71 & 0.61

\\\bottomrule
\end{tabular}
}
\caption{Ablation study results. ``\emph{w/o} $\operatorname{sim} > \ell$'' ablates the textual similarity constraint when constructing the candidate sets, while 
``\emph{w/o} $p_{\text{MLM}} > k$'' ablates the masked language model probability constraint.
}
\label{tab:abl}
\end{table}

\footnotetext{
\Merge perturbation can only merge noun phrases, extracted by the NLTK toolkit(\url{https://www.nltk.org/}).
We find that this helps produce more grammatical outputs.
}


\begin{table}[t]
\centering
\setlength{\tabcolsep}{4.5pt}
\small{
\begin{tabular}{@{} l@{\hskip 0.1mm}rrrrr @{}}
\toprule

\textbf{MLM}  &
\textbf{A-rate}$\uparrow$ & \textbf{Mod}$\downarrow$ & \textbf{PPL}$\downarrow$ & \textbf{Sim}$\uparrow$ & \textbf{Speed}$\uparrow$ \\
\midrule

 $\text{RoBERTa}_{\text{distill}}$ & 79.1 & \textbf{6.1} & \textbf{86.0} & \textbf{0.76} & \textbf{0.14}\\
 $\text{RoBERTa}_{\text{base}}$ & \textbf{79.3} & 6.3 & 88.9 & 0.75 & 0.07\\
 $\text{BERT}_{\text{base}}$ & 78.4 & 8.3 & 95.2 & 0.71 & 0.06
\\\bottomrule
\end{tabular}
}
\caption{Results of \ourmodel implemented with different masked language models (MLM). \textbf{Speed} is measured by number of processed samples per second. 
}
\label{tab:abl_mlm}
\end{table}

We ablate each component of \ourmodel to study its effectiveness.
We evaluate on the 1,000 randomly selected AG news instances~(\S\ref{sec:data}).
The results are summarized in Table~\ref{tab:abl}.

We first investigate the performance of three perturbations when applied individually.
Among three editing strategies, using \textsc{InsertOnly} achieves the best performance,
with \textsc{ReplaceOnly} coming a close second.
\textsc{MergeOnly} underperforms the other two, partly because the attacks are restricted to bigram noun phrases~(\S\ref{sec:setup}). 
Combining all three perturbations, \ourmodel achieves the best performance with the least modifications.

To examine the efficiency of attacking order, we compare \textsc{ReplaceOnly} against BERTAttack. 
Notably, \textsc{ReplaceOnly} outperforms BERTAttack across the board. This is presumably because BERTAttack does not take into account the tokens to be infilled when selecting the attack positions.


We now turn to the two constraints imposed when constructing the candidate token set.
Perhaps not surprisingly, ablating the textual similarity constraint (\emph{w/o} $\operatorname{sim} > l$)
decreases textual similarity performance, but increases other aspects.
Ablating the masked language model yields a better success rate, but much worse perplexity, grammaticality, and textual similarity.

Finally, we compare \ourmodel implemented with different masked language models. Table~\ref{tab:abl_mlm} summarizes the results.
Overall, distilled RoBERTa achieves the fastest speed without losing performance.
Since the victim model is based on BERT, we conjecture that it is less efficient to attack a model using its own information.

\begin{table}[t]
\centering
\setlength{\tabcolsep}{0.0pt}
\small{
\begin{tabular}{@{} lll @{}}
\toprule

\begin{minipage}{0.24\linewidth}
    \textbf{\Replace}
\end{minipage} &
\begin{minipage}{0.39\linewidth}
    \textbf{\Insert} 
\end{minipage} &
\begin{minipage}{0.36\linewidth}
   \textbf{\Merge}
\end{minipage}
\\\midrule

\begin{minipage}{0.24\linewidth}
    \emph{NOUN: 64\%\\ADJ: 17\%\\VERB: 7\%}
\end{minipage}
 &
\begin{minipage}{0.39\linewidth}
   \emph{(NOUN, NOUN): 12\%\\
         (ADJ, NOUN): 10\%\\
         (NOUN, VERB): 9\%}
\end{minipage} 
&
\begin{minipage}{0.36\linewidth}
   \emph{ADJ-NOUN: 31\%\\
         NOUN-NOUN: 22\%\\
         DT-NOUN:  12\%}
\end{minipage}

\\\midrule

\multicolumn{3}{l}{
\begin{minipage}{1\linewidth} 
  \textbf{Context:} ... Amit Yoran, the government's \textsl{cybersecurity} chief, abruptly resigned yesterday after a year ...
\end{minipage}
}\\\noalign{\smallskip}

\multicolumn{3}{l}{
\begin{minipage}{1\linewidth} 
  \textbf{Replace}: \underline{cybersecurity} $\leftarrow$ \emph{\{security, surveillance, cryptography, intelligence, encryption ...\}}
\end{minipage}
}\\\noalign{\smallskip}

\multicolumn{3}{l}{
\begin{minipage}{1\linewidth} 
\textbf{Insert}: cybersecurity \underline{\hskip 5mm} chief $\leftarrow$ \emph{\{technology, defense, intelligence, program, project ...\}}
\end{minipage}
}\\\noalign{\smallskip}

\multicolumn{3}{l}{
\begin{minipage}{1\linewidth} 
  \textbf{Merge}: \underline{cybersecurity chief} $\leftarrow$ \emph{\{chief, consultant, administrator, scientist, secretary ...\}}
\end{minipage}
}\\\bottomrule
\end{tabular}
}
\caption{\textbf{Top}: Top-3 POS tags (or POS tag bigrams) and their percentages for each perturbation type.
$(a,b)$: insert a token between $a$ and $b$.
$a$-$b$: merge $a$ and $b$ into a token. 
\textbf{Bottom}: An AG news sample,
where \ourmodel perturbs token ``\textsl{cybersecurity}.'' 
TextFooler is unable to attack this token since it is out of its vocabularies.}
\label{tab:perturb_stats}
\end{table}

\subsection{Perturbations by Part-of-speech Tags}\label{ana:pos}
In this section, we break down the adversarial attacks by part-of-speech (POS) tags in AG News dataset.
We find that most of the adversarial attacks happen to nouns or noun phrases. Presumably, in many topic classification datasets, the prediction heavily relies on some characteristic  noun words/phrases.
As shown in Table~\ref{tab:perturb_stats}, 64\% of the \Replace actions are applied to nouns.
\Insert actions tend to insert tokens into noun phrase bigram: two of the most frequent POS bigrams are noun phrases.
In fact, around 48\% of the \Insert actions are applied to noun phrases.
This also justifies our choice of only applying \Merge to noun phrases.

\subsection{Adversarial Training} \label{ana:adv_training}
This section explores \ourmodel's potential in improving downstream models' accuracy and robustness.
Following~\citet{tsipras2018robustness},
we use \ourmodel to generate adversarial examples for AG news training instances,
and include them as additional training data.
We consider two settings:
 training with (1) full training data and full adversarial data
and (2) 10\% randomly-sampled training data and its adversarial data, to simulate the low-resource scenario.
For both settings, we compare a BERT-based MLP classifier
and a TextCNN (\citealp{kim2014convolutional}) classifier without any pretrained embedding.

\textit{Whether adversarial examples,
as data augmentation, can help achieve better test accuracy?}
As shown in Table~\ref{tab:adv_train},
when the full training data is available,
adversarial training slightly \emph{decreases} the test accuracy by 0.2\% and 0.5\% respectively.
This aligns with previous observations~\citep{jia2019certified}.
Interestingly, under the low-data scenario with adversarial training,
BERT-based classifier has no accuracy drop, and TextCNN achieves a 2.0\% absolute improvement.
This suggests that a model with less capacity can benefit more from silver data.

\textit{Does adversarial training help the models defend against adversarial attacks?}
To evaluate this, we use \ourmodel to attack the classifiers trained with and without adversarial examples.\footnote{
    In preliminary experiments, we found that
    it is more difficult to use other models to attack a victim model trained with the adversarial examples generated by \ourmodel,
    than to use \ourmodel itself.
}
A higher success rate and fewer modifications indicate a victim classifier is more vulnerable to adversarial attacks.
As shown in Table~\ref{tab:adv_train},
in 3 out of the 4 cases,
adversarial training helps to decrease the attack success rate by more than 10.3\%,
and to increase the number of modifications needed by more than 0.8.
The only exception is the TextCNN model trained with 10\% data.
A possible reason can be that it is trained with few data and thus generalizes less well.

These results suggest that \ourmodel can be used to improve downstream models' robustness,
with a negligible accuracy drop.

\begin{table}[t]
\centering
\small{
\begin{tabular}{@{} lrrr @{}}
\toprule

\textbf{Victim Model}  &
\textbf{Acc}$\uparrow$ & \textbf{A-rate}$\downarrow$ & \textbf{Mod}$\uparrow$
\\\midrule

BERT (100\% data) & 95.0 & 79.1 & 6.1 \\
\quad+ 100\% adversarial & -0.2 & -18.0 & +5.1 \\
\midrule

TextCNN (100\% data) & 91.2 & 92.7 & 5.0 \\
\quad+ 100\% adversarial & -0.5 & -10.3 & +0.8\\

\midrule
\midrule

BERT (10\% data) & 92.5 & 96.1 & 5.4 \\
\quad+ 10\% adversarial & +0.0 & -12.3 & +7.6 \\
\midrule

TextCNN (10\% data) & 83.6 & 99.0 & 5.6 \\
\quad+ 10\% adversarial & +2.0 & -3.5 & +0.3

\\\bottomrule
\end{tabular}
}
\caption{Adversarial training results on AG news test set. ``Acc'' indicates accuracy.}
\label{tab:adv_train}
\end{table}
\section{Related Work}


\paragraph{Textual adversarial attack.}
An increasing amount of effort is being devoted to generating better textual adversarial examples with various attack models.
Character-based models~\interalia{liang2017deep,ebrahimi2018hotflip,li2018textbugger,gao2018black} use misspellings to attack the victim systems; however, these attacks can often be defended by a spell checker~\citep{pruthi2019combating,zhou2019learning,jones2020robust}.
Many sentence-level models~\interalia{iyyer2018adversarial,wang2020t3,zou2019reinforced} have been developed to introduce more sophisticated token/phrase perturbations. These, however, 
generally have difficulty maintaining semantic similarity with original inputs~\citep{zhang2020adversarial}. 
Recent word-level models explore synonym substitution rules to enhance semantic meaning preservation ~\interalia{alzantot2018generating,jin2019bert,ren2019generating,zhang2019generating,zang2020word}. 
Our work differs in that \ourmodel uses three contextualized perturbations that produces more fluent and grammatical outputs. 

\paragraph{Text generation with BERT.} 
Generation with masked language models has been widely studied in various natural language tasks, ranging from lexical substitution~\interalia{wu2019conditional,zhou2019bert,qiang2019simple,wu2019mask} to non-autoregressive generation~\interalia{gu2017non,lee2018deterministic,ghazvininejad2019mask,wang2019bert,ma2019flowseq,sun2019fast,ren2020study, zhang2020pointer}. 


\section{Conclusion}

We have presented \ourmodel, a contextualized adversarial example generation model for text. 
It uses contextualized knowledge from pretrained masked language models,
and can generate adversarial examples that are natural, fluent and grammatical. 
With three contextualized perturbation patterns, \Replace, \Insert and \Merge in our arsenal, \ourmodel can produce outputs of varied lengths and achieves a higher attack success rate than baselines and with fewer edits. Human evaluation shows significant advantages of \ourmodel in terms of textual similarity, fluency and grammaticality. We release our code and models at \url{https://github.com/cookielee77/CLARE}.

\section*{Acknowledgments}
We would like to thank the reviewers for their constructive comments. 
We thank NVIDIA Corporation for the donation of the GPU used for this research. We also thank Tongshuang Wu, Guoyin Wang and Shuhuai Ren for their helpful discussions and feedback. 

\bibliographystyle{acl_natbib}
\bibliography{naacl2020}

\clearpage
\newpage
\appendix
\section{Appendix}

\subsection{Additional Experiment Details}\label{sec:app_details}

\paragraph{Model Implementation.}
All pretrained models and victim models based on RoBERTa and BERT$_{\text{base}}$ are implemented with Hugging Face transformers\footnote{\url{https://github.com/huggingface/transformers}}~\citep{Wolf2019HuggingFacesTS} based on PyTorch~\citep{paszke2019pytorch}. RoBERTa$_{\text{distill}}$, RoBERTa$_{\text{base}}$ and uncase BERT$_{\text{base}}$ models have 82M, 125M and 110M parameters, respectively. We use $\text{RoBERTa}_{\text{distill}}$ as our main backbone for fast inference purpose. 
TextFooler\footnote{\url{https://github.com/jind11/TextFooler}} and BERTAttack\footnote{\url{https://github.com/LinyangLee/BERT-Attack}} are built with their open source implementation provided by the authors. In the implementation of TextFooler+LM, we use small sized GPT-2 language model~\citep{radford2019language} to further select those candidate tokens that have top $20\%$ perplexity in the candidate token set. In the adversarial training (\secref{ana:adv_training}), the small TextCNN victim model~\citep{kim2014convolutional} has 128 embedding size and $100$ filters for $3, 4, 5$ window size with $0.5$ dropout, resulting in 7M parameters. 

During the implementation of \emph{w/o} $p_{\text{MLM}} > k$ in the ablation study (\secref{ana:abl}), we randomly sample 200 tokens and then apply the similarity constraint to construct candidate set, as exhausting the vocabulary is computationally expensive.

\paragraph{Evaluation Metric.}
The similarity function $\operatorname{sim}$ builds on the universal sentence encoder (USE;~\citealp{cer2018universal}) to measure a \emph{local} similarity at the perturbation position with window size 15 between the original input and its adversary. \emph{All baselines} are equipped this $\operatorname{sim}$ when constructing the candidate vocabulary. The evaluation metric \textbf{Sim} uses USE to calculate a \emph{global} similarity between two texts. These procedures are typically following~\citet{jin2019bert}.  We mostly rely on human evaluation (\secref{sec:exp_res}) to conclude the significant advantage of preserving textual similarity on \ourmodel compared with TextFooler.

\paragraph{Data Processing.}
When processing the data, we keep all punctuation in texts for both victim model training and attacking. This differs the pre-processing setting in TextFooler~\citep{jin2019bert} as we empirically found that removing punctuation makes the victim model vulnerable. Since GLUE benchmark~\citep{wang2019glue} does not provide the label for test set, we instead use its dev set as the the test set for the included datasets (MNLI, QNLI, QQP, MRPC, SST-2) in the evaluation. For the sentence-pair tasks (e.g., MNLI, QNLI, QQP, MRPC), we attack the longer one excluding the tokens appearing in both sentences. This is because inference tasks usually require entailed data to have the same keywords, e.g., numbers, name entities, etc. All experiments are conducted on one Nvidia GTX 1080Ti GPU.

\subsection{Additional Results}\label{sec:app_res}

We include the results of DBpedia ontology dataset (\textbf{DBpedia};~\citealp{zhang2015character}, 
Stanford sentiment treebank (\textbf{SST-2};~\citealp{socher2013recursive}),
Microsoft Research Paraphrase Corpus (\textbf{MRPC};~\citealp{dolan2005automatically}), and Quora Question Pairs (\textbf{QQP}) from the GLUE benchmark in this section. Table~\ref{tab:app_dataset} summarizes come statistics of these datasets. 
The results of different models on these datasets are summarized Table~\ref{tab:app_res}. Compared with all baselines, \ourmodel achieves the best performance on attack success rate, perplexity, grammaticality, and similarity. It is consistent with our observation in \secref{sec:exp_res}.

\begin{table}[t]
\centering
\setlength{\tabcolsep}{4.0pt}
\small{
\begin{tabular}{@{} lccrrr @{}}
\toprule

\textbf{Dataset}  & \textbf{Avg. Length} & \textbf{\# Classes} &
\textbf{Train} & \textbf{Test} & \textbf{Acc}
\\\midrule

SST-2 & 10 & 2 & 67K & 0.9K & 92.3\% \\
DBpedia & 55 & 14 & 560K & 70K & 99.3\%  \\
\midrule

QQP & 13/13 & 2 & 363K & 40K & 91.4\%\\
MRPC & 23/23 & 2 & 3.6K & 1.7K & 81.4\%
\\\bottomrule
\end{tabular}
}
\caption{Some statistics of datasets. 
The last column indicates the victim model's accuracy on the original test set \emph{without} adversarial attack.}
\label{tab:app_dataset}
\end{table}



\begin{table*}[t]
\centering
\setlength{\tabcolsep}{6pt}
\small{
\begin{tabular}{@{} l rrrrr m{3pt} rrrrr @{}}
\toprule
{} & \multicolumn{5}{c}{SST-2 (PPL = 99.5)} && \multicolumn{5}{c}{DBpedia (PPL = 37.3)}
\\\midrule
\textbf{Model}  &
\textbf{A-rate}$\uparrow$ & \textbf{Mod}$\downarrow$ & \textbf{PPL}$\downarrow$ & \textbf{GErr}$\downarrow$ & \textbf{Sim}$\uparrow$ & &
\textbf{A-rate}$\uparrow$ & \textbf{Mod}$\downarrow$ & \textbf{PPL}$\downarrow$ & \textbf{GErr}$\downarrow$ & \textbf{Sim}$\uparrow$
\\\midrule
 TextFooler  &
 89.8 &  14.9 &  227.7 &  0.53 &  0.69 &&
 56.2 &  24.9 &  182.5 &  1.88 &  0.68
 \\
 \quad + LM  &
 51.7 &  18.3 &  137.5 &  0.50 &  0.69 &&
 20.1 &  22.4 &  84.0 &  1.22 &  0.70
 \\
  BERTAttack  &
 87.8 &  8.1 &  142.9 &  0.03 &  0.67 &&
 60.7 &  9.1 &  57.8 &  0.20 &  0.69
 \\
 \textsc{Clare}  &
 \textbf{97.8} &  \textbf{7.5} &  \textbf{137.4} &  \textbf{0.01} &  \textbf{0.75} &&
\textbf{65.8} &  \textbf{7.0} &  \textbf{53.3} &  \textbf{-0.03} &  \textbf{0.73}
\\\midrule\midrule

{} & \multicolumn{5}{c}{QQP (PPL = 56.2)} && \multicolumn{5}{c}{MRPC (PPL = 42.9)}
\\\midrule
\textbf{Model}  &
\textbf{A-rate}$\uparrow$ & \textbf{Mod}$\downarrow$ & \textbf{PPL}$\downarrow$ & \textbf{GErr}$\downarrow$ & \textbf{Sim}$\uparrow$ & &
\textbf{A-rate}$\uparrow$ & \textbf{Mod}$\downarrow$ & \textbf{PPL}$\downarrow$ & \textbf{GErr}$\downarrow$ & \textbf{Sim}$\uparrow$
\\\midrule
 TextFooler  &
 16.2 &  12.7 &  145.2 &  0.61 &  0.74 &&
 24.5 &  10.6 &  118.8 &  0.35 &  0.75
 \\
 \quad + LM  &
 7.8 &  12.9 &  78.8 &  0.21 &  0.77 &&
 12.9 &  9.5 &  71.0 &  0.29 &  0.79
 \\
  BERTAttack  &
 24.2 &  11.3 &  78.0 &  0.25 &  0.71 &&
 29.7 &  13.5 &  74.6 &  0.05 &  0.79
 \\
 \textsc{Clare}  &
\textbf{27.7} &  \textbf{10.2} &  \textbf{74.8} &  \textbf{0.14} &  \textbf{0.76} &&
\textbf{34.8} &  \textbf{9.1} &  \textbf{69.5} &  \textbf{0.02} &  \textbf{0.83}
\\\bottomrule
\end{tabular}
}
\caption{Adversarial example generation performance in
attack success rate (A-rate),
modification rate (Mod),
perplexity (PPL),
number of increased grammar errors (GErr),
and text similarity (Sim).
The perplexity of the original inputs is indicated in parentheses for each dataset.
Bold indicates the best performance on each metric.}
\label{tab:app_res}
\end{table*}










\subsection{Human Evaluation Details}\label{sec:app_human_eval}

For each human evaluation on \textbf{AG News} dataset, we randomly sampled 300 sentences from the test set combining the corresponding adversarial examples from \ourmodel and TextFooler (We only consider sentences can be attacked by both models). 
In order to make the task less abstract, we pair the adversarial examples by the two models, and present them to the participants along with the original input and its gold label.
We ask them which one they prefer in terms of
(1) having more similar a meaning to the original input (similarity), 
and (2) being more fluent and grammatical (fluency and grammaticality).
We also provide them  with a 
neutral option, when the participants consider the two indistinguishable.
Additionally, we ask the participants to annotate the adversarial examples,
and compare their annotations against the gold labels (label consistency).
Higher label consistency indicates the model is better at causing the victim model to make errors while preserving human predictions.

\begin{figure*}[t]
\centering
\includegraphics[width=1.0\linewidth]{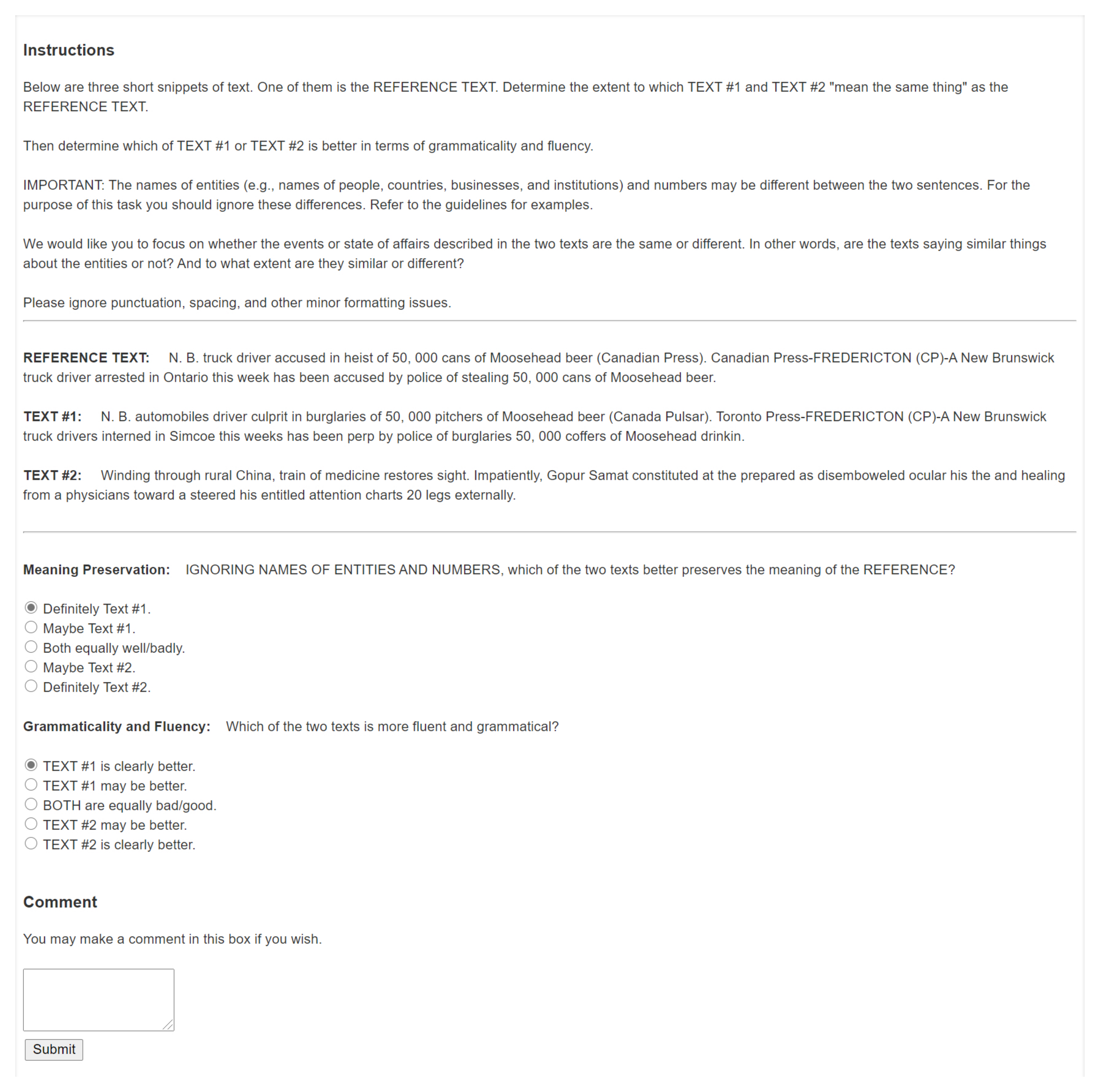}
\caption{Pair-wise comparison in terms of text similarity and fluency \& grammaticality on human evaluation.}
\label{fig:app_human_compare}
\end{figure*}

\begin{figure*}[t]
\centering
\includegraphics[width=1.0\linewidth]{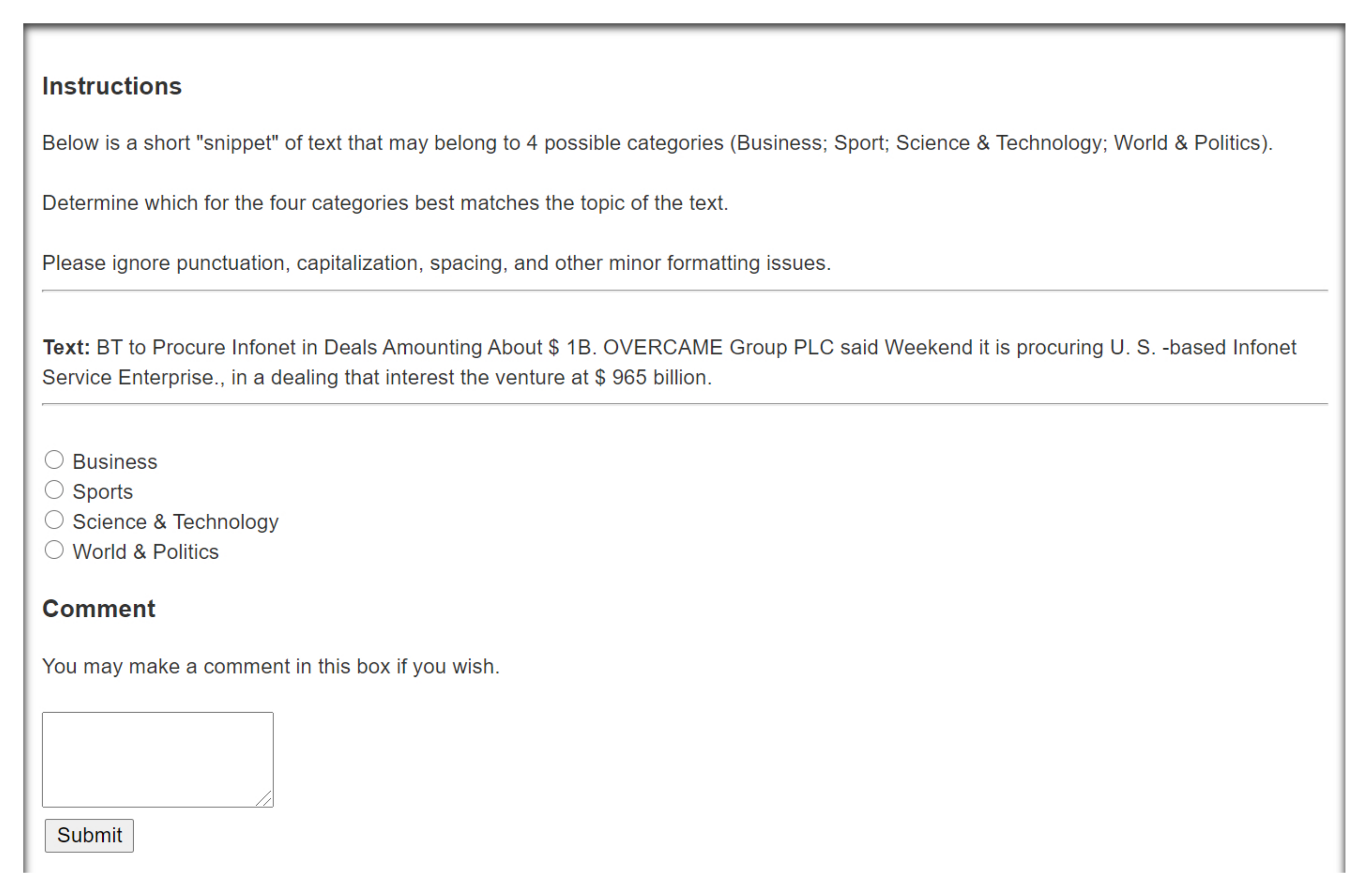}
\caption{Label consistency task on human evaluation.}
\label{fig:app_human_label}
\end{figure*}

Each pair of system outputs was randomly presented to 5 crowd-sourced judges, who indicated their preference for similarity, fluency, and grammaticality using the form shown in Figure~\ref{fig:app_human_compare}. The labelling task is illustrated in Figure~\ref{fig:app_human_label}. To minimize the impact of spamming, we employed the top-ranked 30\% of U.S. workers provided by the crowd-sourcing service. Detailed task descriptions and examples were also provided to guide the judges. We calculate $p$-value based on 95\% confidence intervals by using 10K paired bootstrap replications, implemented using the R Boot statistical package.

\subsection{Qualitative Samples}\label{sec:app_samples}

We include generated adversarial examples by \ourmodel and TextFooler on \textbf{AG News}, \textbf{DBpeida}, \textbf{Yelp}, \textbf{MNLI},  and \textbf{QNLI} datasets in Table~\ref{tab:app_sample1} and Table~\ref{tab:app_sample2}.


\begin{table*}[t]
\small
\begin{tabular}{@{} ll @{}}

\noalign{\smallskip}\Xhline{3\arrayrulewidth}\noalign{\smallskip}


 

 \begin{minipage}{0.1\textwidth}
   \textbf{AG} \\
   (Business)
\end{minipage}
&  
\begin{minipage}{0.9\textwidth}
   TECH BUZZ : Yahoo, Adobe team up for new Web services. Stepping up the battle of online search and services, Yahoo Inc. and Adobe Systems Inc. have joined forces to tap each other's customers and put Web search features into Adobe's popular Acrobat Reader software.
\end{minipage}
 \\\noalign{\smallskip}\hdashline\noalign{\smallskip}

\begin{minipage}{0.1\textwidth}
   TextFooler \\
   (Sci\&Tech)
\end{minipage}
&  
\begin{minipage}{0.9\textwidth}
   TECH BUZZ : Yahoo, Adobe team up for \replacecolor{roman} \replacecolor{Cyberspace} \replacecolor{utilities}. Stepping up the battle of online \replacecolor{locating} and services, Yahoo Inc. and Adobe Systems Inc. have joined forces to tap each other's customers and put Web search features into Adobe's popular Acrobat Reader software.
\end{minipage}
 \\\noalign{\smallskip}\hdashline\noalign{\smallskip}
 
 \begin{minipage}{0.1\textwidth}
   \ourmodel \\
   (Sci\&Tech)
\end{minipage}
&  
\begin{minipage}{0.9\textwidth}
   TECH BUZZ : Yahoo, Adobe team up for new Web \replacecolor{Explorer}. Stepping up the battle of online search and services, Yahoo Inc. and Adobe Systems Inc. have joined forces to tap each other's customers and put Web search features into Adobe's popular Acrobat Reader software.
\end{minipage}
 \\\noalign{\smallskip}\Xhline{3\arrayrulewidth}\noalign{\smallskip}

 \begin{minipage}{0.1\textwidth}
   \textbf{AG} \\
   (Sport)
\end{minipage}
&  
\begin{minipage}{0.9\textwidth}
   Padres Blank Dodgers 3 - 0. LOS ANGELES - Adam Eaton allowed five hits over seven innings for his career - high 10th victory, Brian Giles homered for the second straight game, and the San Diego Padres beat the Los Angeles Dodgers 3 - 0 Thursday night. The NL West - leading Dodgers' lead was cut to 2 1 / 2 games over San Francisco - their smallest since July 31 ...
\end{minipage}
 \\\noalign{\smallskip}\hdashline\noalign{\smallskip}

\begin{minipage}{0.1\textwidth}
   TextFooler \\
   (World)
\end{minipage}
&  
\begin{minipage}{0.9\textwidth}
   Dodger Blank \replacecolor{Yanks} 3 - 0. \replacecolor{Loos} ANGELES - \replacecolor{Adams} \replacecolor{Parades} \replacecolor{enabling} five hits over seven \replacecolor{slugging} for his career - high 10th \replacecolor{victoria}, Brian Giles homered for the second straight \replacecolor{matching}, and the \replacecolor{Tome} \replacecolor{José} \replacecolor{Dodger} beat the Los Angeles \replacecolor{Dodger} 3 - 0 Thursday \replacecolor{blackness}. The NL \replacecolor{Westerner} - \replacecolor{eminent} \replacecolor{Dodger}' lead was cut to 2 1 / 2 games over San \replacecolor{San} - their \replacecolor{tiny} \replacecolor{as} \replacecolor{janvier} 31 ...
\end{minipage}
 \\\noalign{\smallskip}\hdashline\noalign{\smallskip}
 
 \begin{minipage}{0.1\textwidth}
   \ourmodel \\
   (World)
\end{minipage}
&  
\begin{minipage}{0.9\textwidth}
   Padres Blank Dodgers 3 - 0. \mergecolor{Milwaukee} \insertcolor{NEXT} - Adam Eaton allowed five hits over seven innings for his career - high 10th victory, Brian Giles homered for the second straight game, and the San Diego Padres beat the Los Angeles Dodgers 3 - 0 Thursday night. The NL West - leading Dodgers' lead was cut to 2 1 / 2 games over San Francisco - their smallest since July 31 ...
\end{minipage}
 \\\noalign{\smallskip}\Xhline{3\arrayrulewidth}\noalign{\smallskip}

 
 \begin{minipage}{0.1\textwidth}
  \textbf{Yelp} \\
  (Positive)
\end{minipage}
&  
\begin{minipage}{0.9\textwidth}
  The food at this chain has always been consistently good. Our server in downtown ( where we spent New Year's ) was new, but that did not impact our service at all. She was prompt and attentive to our needs.
\end{minipage}
 \\\noalign{\smallskip}\hdashline\noalign{\smallskip}

\begin{minipage}{0.1\textwidth}
  TextFooler \\
  (Negative)
\end{minipage}
&  
\begin{minipage}{0.9\textwidth}
  The food at this chain has always been \replacecolor{necessarily} \replacecolor{ok}. Our server in downtown ( where we spent New Year's ) was new, but that did not impact our service at all. She was \replacecolor{early} and attentive to our needs.
\end{minipage}
 \\\noalign{\smallskip}\hdashline\noalign{\smallskip}
 
 \begin{minipage}{0.1\textwidth}
  \ourmodel \\
  (Negative)
\end{minipage}
&  
\begin{minipage}{0.9\textwidth}
  The food at this chain has always been \insertcolor{looking} consistently good. Our server in downtown ( where we spent New Year's ) was new, but that did not \replacecolor{enhance} our service at all. She was prompt and attentive to our needs.
\end{minipage}
 \\\noalign{\smallskip}\Xhline{3\arrayrulewidth}\noalign{\smallskip}
 
\begin{minipage}{0.1\textwidth}
  \textbf{Yelp} \\
  (Positive)
\end{minipage}
&  
\begin{minipage}{0.9\textwidth}
  The pho broth is actually flavorful and doesn't just taste like hot water with beef and noodles. I usually do take out and the order comes out fast during dinner which should be expected with pho, it's not hard to soak noodles, slice beef and pour broth.
\end{minipage}
 \\\noalign{\smallskip}\hdashline\noalign{\smallskip}

\begin{minipage}{0.1\textwidth}
  TextFooler \\
  (Negative)
\end{minipage}
&  
\begin{minipage}{0.9\textwidth}
  The pho broth is actually flavorful and doesn't just \replacecolor{tasty} like \replacecolor{torrid} \replacecolor{waters} with \replacecolor{slaughter} and \replacecolor{salads}. I \replacecolor{repeatedly} do take out and the order \replacecolor{poses} out fast during dinner which should be expected with pho , it's not \replacecolor{strenuous} to soak noodles, \replacecolor{severing} beef and pour broth.
\end{minipage}
 \\\noalign{\smallskip}\hdashline\noalign{\smallskip}
 
 \begin{minipage}{0.1\textwidth}
  \ourmodel \\
  (Negative)
\end{minipage}
&  
\begin{minipage}{0.9\textwidth}
  The pho broth is actually flavorful and doesn't just taste \insertcolor{bland} like hot water with beef and noodles. I usually do take out and the order comes out \insertcolor{awfully} fast during dinner which should be expected with pho, it's not hard to soak noodles, slice beef and pour broth.
\end{minipage}
 \\\noalign{\smallskip}\Xhline{3\arrayrulewidth}\noalign{\smallskip}


 
 
 \begin{minipage}{0.1\textwidth}
   \textbf{MNLI} \\
   (Neutral)
\end{minipage}
&  
\begin{minipage}{0.9\textwidth}
    \emph{Premise}: Thebes held onto power until the 12th Dynasty, when its first king, Amenemhet Iwho reigned between 1980 1951 b.c. established a capital near Memphis. \\
   \emph{Hypothesis}:  The capital near Memphis lasted only half a century before its inhabitants abandoned it for the next capital.
\end{minipage}
 \\\noalign{\smallskip}\hdashline\noalign{\smallskip}

\begin{minipage}{0.1\textwidth}
   TextFooler \\
   (Contradiction)
\end{minipage}
&  
\begin{minipage}{0.9\textwidth}
   \emph{Premise}: Thebes \replacecolor{apprehended} \replacecolor{pour} \replacecolor{powers} until the 12th \replacecolor{Familial} , when its \replacecolor{earliest} king , Amenemhet Iwho reigned between 1980 1951 \replacecolor{c}.c. established a capital near Memphis . \\
   \emph{Hypothesis}:  The capital near Memphis lasted only half a century before its inhabitants abandoned it for the next capital.
\end{minipage}
 \\\noalign{\smallskip}\hdashline\noalign{\smallskip}
 
 \begin{minipage}{0.1\textwidth}
   \ourmodel \\
   (Contradiction)
\end{minipage}
&  
\begin{minipage}{0.9\textwidth}
   \emph{Premise}: Thebes held onto power until the 12th Dynasty, when its first king, Amenemhet Iwho reigned between 1980 1951 b.c. \insertcolor{thereafter} established a capital near Memphis. \\
   \emph{Hypothesis}:  The capital near Memphis lasted only half a century before its inhabitants abandoned it for the next capital.
\end{minipage}
 \\\noalign{\smallskip}\Xhline{3\arrayrulewidth}\noalign{\smallskip}
 
 
 \begin{minipage}{0.1\textwidth}
   \textbf{MNLI} \\
   (Entailment)
\end{minipage}
&  
\begin{minipage}{0.9\textwidth}
   \emph{Premise}: Hopefully, Wall Street will take voluntary steps to address these issues before it is forced to act.\\
   \emph{Hypothesis}:  Wall Street is facing issues, that need to be addressed.
\end{minipage}
 \\\noalign{\smallskip}\hdashline\noalign{\smallskip}

\begin{minipage}{0.1\textwidth}
   TextFooler\\
   (Neutral)
\end{minipage}
&  
\begin{minipage}{0.9\textwidth}
   \emph{Premise}: Hopefully, Wall Street will take voluntary steps to \replacecolor{treatment} these issues before it is forced to act.\\
   \emph{Hypothesis}:  Wall Street is facing issues, that need to be addressed.
\end{minipage}
 \\\noalign{\smallskip}\hdashline\noalign{\smallskip}
 
 \begin{minipage}{0.1\textwidth}
   \ourmodel\\
   (Neutral)
\end{minipage}
&  
\begin{minipage}{0.9\textwidth}
   \emph{Premise}: Hopefully, Wall Street will take voluntary steps to \replacecolor{eliminate} these issues before it is forced to act.\\
   \emph{Hypothesis}:  Wall Street is facing issues, that need to be addressed.
\end{minipage}
\\\noalign{\smallskip}\Xhline{3\arrayrulewidth}\noalign{\smallskip}
\end{tabular}
\caption{Adversarial examples produced by different models. The gold label of the original is shown below the (bolded) dataset name. \replacecolor{\textbf{Replace}}, \insertcolor{\textbf{Insert}} and \mergecolor{\textbf{Merge}} are highlighted in \replacecolor{italic red}, \insertcolor{bold blue} and \mergecolor{sans serif yellow}, respectively. (Best viewed in color).}
\label{tab:app_sample1}
\end{table*}


\begin{table*}[t]
\small
\begin{tabular}{@{} ll @{}}

\\\noalign{\smallskip}\Xhline{3\arrayrulewidth}\noalign{\smallskip}

 \begin{minipage}{0.1\textwidth}
   \textbf{QNLI} \\
   (Entailment)
\end{minipage}
&  
\begin{minipage}{0.9\textwidth}
   \emph{Premise}: Who overturned the Taft Vale judgement ?\\
   \emph{Hypothesis}:  One of the first acts of the new Liberal Government was to reverse the Taff Vale judgement.
\end{minipage}
 \\\noalign{\smallskip}\hdashline\noalign{\smallskip}

\begin{minipage}{0.1\textwidth}
   TextFooler\\
   (Not-\\Entailment)
\end{minipage}
&  
\begin{minipage}{0.9\textwidth}
   \emph{Premise}: Who overturned the Taft Vale judgement ?\\
   \emph{Hypothesis}:  One of the first acts of the new Liberal Government was to \replacecolor{invest} the Taff Vale judgement.
\end{minipage}
 \\\noalign{\smallskip}\hdashline\noalign{\smallskip}
 
 \begin{minipage}{0.1\textwidth}
   \ourmodel\\
   (Not-\\Entailment)
\end{minipage}
&  
\begin{minipage}{0.9\textwidth}
   \emph{Premise}: Who overturned the Taft Vale judgement ?\\
   \emph{Hypothesis}:  One of the first acts of the new Liberal \replacecolor{Constitution} was to reverse the Taff Vale judgement.
\end{minipage}
\\\noalign{\smallskip}\Xhline{3\arrayrulewidth}\noalign{\smallskip}


 \begin{minipage}{0.1\textwidth}
   \textbf{QNLI} \\
   (Entailment)
\end{minipage}
&  
\begin{minipage}{0.9\textwidth}
   \emph{Premise}: What are the software testers aware of ?\\
   \emph{Hypothesis}:  Black-box testing treats the software as a black box, examining functionality without any knowledge of internal implementation, without seeing the source code.
\end{minipage}
 \\\noalign{\smallskip}\hdashline\noalign{\smallskip}

\begin{minipage}{0.1\textwidth}
   TextFooler\\
   (Not-\\Entailment)
\end{minipage}
&  
\begin{minipage}{0.9\textwidth}
   \emph{Premise}: What are the software testers aware of ?\\
   \emph{Hypothesis}: Black-\replacecolor{boxes} testing \replacecolor{administers} the software as a black box, \replacecolor{investigating} \replacecolor{functions} \replacecolor{unless} any knowledge of internal \replacecolor{fulfil}, \replacecolor{unless} seeing the \replacecolor{wellspring} code.
\end{minipage}
 \\\noalign{\smallskip}\hdashline\noalign{\smallskip}
 
 \begin{minipage}{0.1\textwidth}
   \ourmodel\\
   (Not-\\Entailment)
\end{minipage}
&  
\begin{minipage}{0.9\textwidth}
   \emph{Premise}: What are the software testers aware of ?\\
   \emph{Hypothesis}:  Black-box testing treats the software as a black box, examining functionality without \replacecolor{awareness} of internal implementation, without seeing the source code.
\end{minipage}
\\\noalign{\smallskip}\Xhline{3\arrayrulewidth}\noalign{\smallskip}


 \begin{minipage}{0.1\textwidth}
   \textbf{DBpedia} \\
   (Transportation)
\end{minipage}
&  
\begin{minipage}{0.9\textwidth}
   Honda Crossroad. The Honda Crossroad refers to two specific types of SUVs made by Honda. One of them is a rebadged Land Rover Discovery Series I SUV while the other is a completely different vehicle introduced in 2008.
\end{minipage}
 \\\noalign{\smallskip}\hdashline\noalign{\smallskip}

\begin{minipage}{0.1\textwidth}
   TextFooler\\
   (Album)
\end{minipage}
&  
\begin{minipage}{0.9\textwidth}
   \replacecolor{Suzuki} \replacecolor{Junctions}. The \replacecolor{Suzuki} Crossroad refers to \replacecolor{three} \replacecolor{accurate} \replacecolor{typing} of \replacecolor{prius} \replacecolor{posed} by\replacecolor{Isuzu}. One of them is a rebadged Land Rover \replacecolor{Identify} Series I \replacecolor{LEXUS} while the other is a completely different vehicle introduced in 2008.
\end{minipage}
 \\\noalign{\smallskip}\hdashline\noalign{\smallskip}
 
 \begin{minipage}{0.1\textwidth}
   \ourmodel\\
   (Company)
\end{minipage}
&  
\begin{minipage}{0.9\textwidth}
   Honda Crossroad. The Honda Crossroad refers to two specific \replacecolor{manufacturers} of SUVs made by Honda. One of them is a rebadged Land Rover Discovery Series I SUV while the other is a completely different vehicle introduced in 2008.
\end{minipage}
\\\noalign{\smallskip}\Xhline{3\arrayrulewidth}\noalign{\smallskip}


 \begin{minipage}{0.1\textwidth}
   \textbf{DBpedia} \\
   (Company)
\end{minipage}
&  
\begin{minipage}{0.9\textwidth}
   Yellow Rat Bastard. Yellow Rat Bastard is the flagship establishment in a chain of New York City retail clothing stores owned by Henry Ishay. It specializes in hip - hop-and alternative - style clothing and shoes.
\end{minipage}
 \\\noalign{\smallskip}\hdashline\noalign{\smallskip}

\begin{minipage}{0.1\textwidth}
   TextFooler\\
   (Building)
\end{minipage}
&  
\begin{minipage}{0.9\textwidth}
  \replacecolor{Yellowish} \replacecolor{Rats} \replacecolor{Schmuck} . \replacecolor{Yellowish} \replacecolor{Rats} \replacecolor{Dickwad} is the flagship \replacecolor{establishments} in a \replacecolor{chains} of New York City retail \replacecolor{uniforms} stores owned by \replacecolor{Henrik} Ishay . It \replacecolor{specialize} in hip - hop-and alternative - style \replacecolor{laundry} and \replacecolor{sneakers}.
\end{minipage}
 \\\noalign{\smallskip}\hdashline\noalign{\smallskip}
 
 \begin{minipage}{0.1\textwidth}
   \ourmodel\\
   (Building)
\end{minipage}
&  
\begin{minipage}{0.9\textwidth}
   Yellow Rat Bastard. Yellow Rat Bastard \insertcolor{Mall} is the flagship establishment in a chain of New York City retail clothing stores owned by Henry Ishay. It specializes in hip - hop-and alternative - style clothing and shoes.
\end{minipage}
\\\noalign{\smallskip}\Xhline{3\arrayrulewidth}\noalign{\smallskip}


 \begin{minipage}{0.1\textwidth}
  \textbf{MRPC} \\
  (Not \\ Paraphrase)
\end{minipage}
&  
\begin{minipage}{0.9\textwidth}
  \emph{Premise}: The Americas market will decline 2.1 percent to \$30.6 billion in 2003, and then grow 15.7 percent to \$35.4 billion in 2004.\\
  \emph{Hypothesis}: The US chip market is expected to decline 2.1 percent this year, then grow 15.7 percent in 2004.
\end{minipage}
 \\\noalign{\smallskip}\hdashline\noalign{\smallskip}

\begin{minipage}{0.1\textwidth}
  TextFooler\\
  (Paraphrase)
\end{minipage}
&  
\begin{minipage}{0.9\textwidth}
  \emph{Premise}: The Americas market will decline 2.1 percent to \$30.6 billion in 2003, and then grow 15.7 percent to \$35.4 billion in 2004.\\
  \emph{Hypothesis}: The US chip market is \replacecolor{prescribed} to decline 2.1 percent this year, then grow 15.7 percent in 2004.
\end{minipage}
 \\\noalign{\smallskip}\hdashline\noalign{\smallskip}
 
 \begin{minipage}{0.1\textwidth}
  \ourmodel\\
  (Paraphrase)
\end{minipage}
&  
\begin{minipage}{0.9\textwidth}
  \emph{Premise}: The Americas market will decline 2.1 percent to \$30.6 billion in 2003, and then grow 15.7 percent to \$35.4 billion in 2004.\\
  \emph{Hypothesis}: The US chip market is expected to decline 2.1 percent this year, then grow 15.7 percent in 2004 \insertcolor{yr}.
\end{minipage}
\\\noalign{\smallskip}\Xhline{3\arrayrulewidth}\noalign{\smallskip}


 \begin{minipage}{0.1\textwidth}
  \textbf{MRPC} \\
  (Paraphrase)
\end{minipage}
&  
\begin{minipage}{0.9\textwidth}
  \emph{Premise}: The Securities and Exchange Commission filed a civil fraud suit against the teen in Boston.\\
  \emph{Hypothesis}: The Securities and Exchange Commission brought a related civil case on Thursday.
\end{minipage}
 \\\noalign{\smallskip}\hdashline\noalign{\smallskip}

\begin{minipage}{0.1\textwidth}
  TextFooler\\
  (Not \\ Paraphrase)
\end{minipage}
&  
\begin{minipage}{0.9\textwidth}
  \emph{Premise}: The Securities and Exchange Commission filed a civil fraud suit against the teen in Boston.\\
  \emph{Hypothesis}: The Securities and Exchange Commission brought a \replacecolor{connect} civil case on \replacecolor{Yesterday}.
\end{minipage}
 \\\noalign{\smallskip}\hdashline\noalign{\smallskip}
 
 \begin{minipage}{0.1\textwidth}
  \ourmodel\\
  (Not \\ Paraphrase)
\end{minipage}
&  
\begin{minipage}{0.9\textwidth}
  \emph{Premise}: The Securities and Exchange Commission filed a civil fraud suit against the teen in Boston.\\
  \emph{Hypothesis}: The Securities and Exchange Commission brought a \replacecolor{Massachusetts} civil \replacecolor{lawsuit} on Thursday.
\end{minipage}
\\\noalign{\smallskip}\Xhline{3\arrayrulewidth}\noalign{\smallskip}
\end{tabular}
\caption{Adversarial examples produced by different models. The gold label of the original is shown below the (bolded) dataset name. \replacecolor{\textbf{Replace}}, \insertcolor{\textbf{Insert}} and \mergecolor{\textbf{Merge}} are highlighted in \replacecolor{italic red}, \insertcolor{bold blue} and \mergecolor{sans serif yellow}, respectively. (Best viewed in color).}
\label{tab:app_sample2}
\end{table*}

\clearpage
\newpage

\end{document}